\documentclass[sigconf,nonacm]{acmart} 

\AtBeginDocument{%
  }

\setcopyright{acmlicensed}
\copyrightyear{2026}
\acmYear{2026}
\acmDOI{XXXXXXX.XXXXXXX}
\acmConference[CIKM '26]{The 35th ACM International Conference on Information and Knowledge Management}{Nov 07--11,
  2026}{Rome, ITALY}
\acmISBN{978-1-4503-XXXX-X/2018/06}

\newcommand{\method}{\textsf{HADES}}

\usepackage{xcolor}
\usepackage{balance}
\usepackage{mathtools}
\usepackage{url}
\usepackage{amsfonts}
\usepackage{algorithmic}
\usepackage{textcomp}
\usepackage{algorithm}
\usepackage{multirow}
\usepackage{color}
\usepackage{subcaption}
\usepackage{enumitem}

\usepackage{subcaption}
\usepackage{pgfplots}
\usetikzlibrary{pgfplots.groupplots}
\usetikzlibrary{patterns}
\usetikzlibrary{backgrounds,calc}
\def\addlegendimage{\csname pgfplots@addlegendimage\endcsname}

\begin{document}


\title[Heterophily-Aware Adaptive Knowledge Distillation for Hypergraph Neural Networks]{Heterophily-Aware Adaptive Knowledge Distillation for Hypergraph Neural Networks}

\author{Joohee Cho}
\affiliation{%
  \institution{Chung-Ang University}
  \city{Seoul}
  \country{South Korea}}
\email{joohee2008@cau.ac.kr}

\author{David Yoon Suk Kang}
\affiliation{%
  \institution{Chungbuk National University}
  \city{Cheongju}
  \country{South Korea}}
\email{dyskang@cbnu.ac.kr}

\author{Yunyong Ko}
\authornote{Corresponding author}
\affiliation{%
  \institution{Chung-Ang University}
  \city{Seoul}
  \country{South Korea}}
\email{yyko@cau.ac.kr}


\begin{abstract}
\textit{Hypergraph knowledge distillation} aims to retain the predictive performance of a hypergraph neural network (HNN) teacher while reducing inference costs through a lightweight student model. 
In this work, we observe that HNNs exhibit substantially lower prediction performance on \textit{heterophilic} nodes connected through semantically diverse hyperedges, indicating that the reliability of teacher knowledge varies across nodes. 
Motivated by this observation, we propose \textbf{{\method}}, a heterophily-aware adaptive distillation method for hypergraph neural networks. 
{\method} quantifies node heterophily and leverages it as an estimate of teacher reliability to modulate the transfer of teacher knowledge during distillation. 
Experimental results on real-world hypergraphs demonstrate that {\method} consistently improves student performance across different HNN teachers and distillation objectives. 
In many cases, the resulting student models surpass the predictive performance of their teachers while achieving up to $12.3\times$ faster inference.

\end{abstract}




\maketitle

\section{Introduction}\label{sec:intro}
\textit{Hypergraph neural networks} (HNNs) have achieved superior performance to conventional graph neural networks (GNNs) in a wide range of applications~\cite{ko2023cash,yu2025hygen,ko2025learning,feng2019hypergraph,dong2020hnhn,chien2021you,10.1145/3627673.3679646,10.1145/3583780.3615054,multiway} by effectively modeling high-order relationships among multiple entities. 
However, such expressive power requires high computational cost, limiting the practical deployment of HNNs~\cite{feng2024lighthgnn,forouzandeh2025distillhgnn}. 
To address this issue, a handful of recent studies have explored \textit{hypergraph knowledge distillation}~\cite{xu2026adaptive,feng2024lighthgnn,forouzandeh2025distillhgnn}, 
which transfers the knowledge learned by a powerful HNN teacher to a lightweight MLP student. 
Hypergraph knowledge distillation aims to preserve teacher-level predictive performance while enabling substantially faster inference.

Through hypergraph knowledge distillation, 
a lightweight MLP student can learn the high-order relational knowledge captured by an HNN teacher without directly modeling the hypergraph structure.
Consequently, nodes sharing similar semantics (e.g., class labels) tend to have similar representations, 
while semantically different nodes are separated in the learned representation space~\cite{feng2024lighthgnn,chien2021you,HNHN2020,BAI2021107637}. 
Therefore, the effectiveness of hypergraph distillation largely depends on the reliability of the transferred teacher knowledge.

Then, a natural question to raise is ``\textit{Is the transferred teacher knowledge equally reliable across all nodes?}"
Despite its importance, this question has received little attention in existing hypergraph distillation studies. 
To answer this question, we conduct a preliminary analysis of the reliability of HNN teacher knowledge.
We first hypothesize that, compared with \textit{homophilic} nodes connected mainly to same-label neighbors~\cite{10.5555/3495724.3496377,10.1145/3583780.3615195,10.1145/3627673.3679604,10.1145/3627673.3679622}, 
\textit{heterophilic} nodes connected to many different-label neighbors are more difficult for HNNs to model and therefore may yield less reliable teacher knowledge.

To verify this hypothesis, we first count the number of heterophilic neighbors of each node in real-world hypergraphs and partition nodes into five groups accordingly (G1--G5). 
We then measure the node classification accuracy of two HNN models (HGNN~\cite{feng2019hypergraph} and UniGCN~\cite{huang2021unignn}) and a 3-layer MLP model on each group.

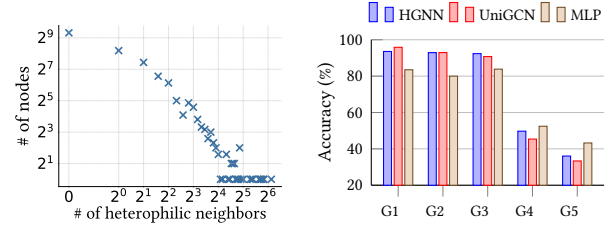
\begin{figure}
    \begin{tabular}{ll}
        \includegraphics[width=0.42\linewidth]{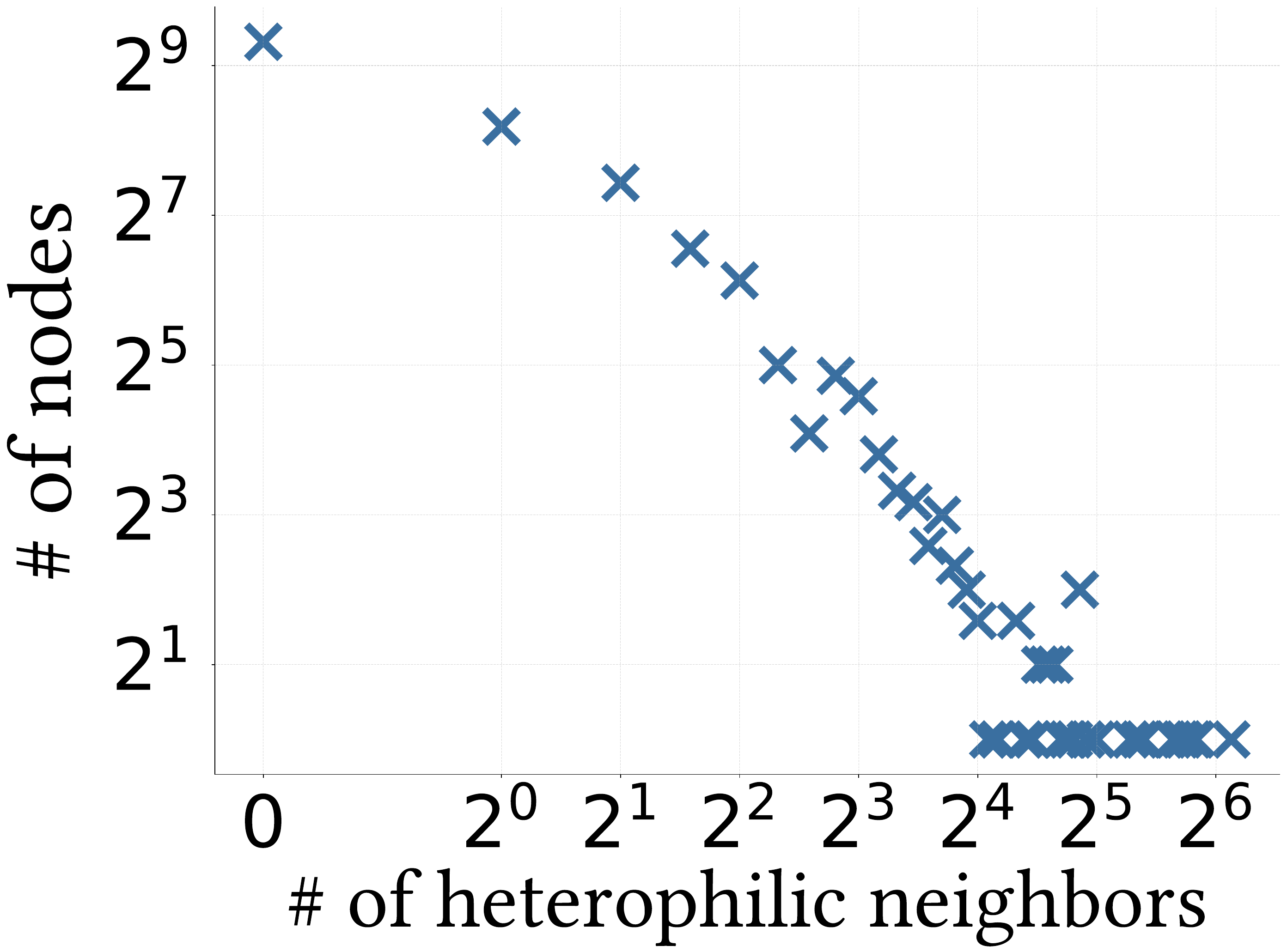}
        & 
        \begin{tikzpicture}	
        \begin{axis}[
            height=3.5cm,
            width=0.55\linewidth,
            axis x line*=center,
            axis y line*=left,
            ymajorgrids=true,
            major grid style={line width=.2pt,draw=gray!50},
            ybar=1pt,
            bar width=3pt,
            enlarge x limits=0.15,
            ylabel=Accuracy (\%),
            ylabel style={yshift=-18pt, font=\footnotesize},
            yticklabel style={font=\scriptsize},
            ymin=20,
            ymax=100,
	        title style={font=\small,yshift=-82pt},
            symbolic x coords={G1,G2,G3,G4,G5},
            xtick=data,
            xticklabels={G1,G2,G3,G4,G5},
            xticklabel style={
                font=\scriptsize,
                anchor=north,
                yshift=0pt,
                xshift=-3pt,
            },
            legend style={at={(0.5,1.3)}, anchor=north, legend columns=-1, font=\scriptsize, draw=none},
        ]
        
        \addplot
        coordinates {
            (G1,93.56)
            (G2,92.91)
            (G3,92.37)
            (G4,49.72)
            (G5,36.05)
        };
        \addplot
        coordinates {
            (G1,95.85)
            (G2,92.95)
            (G3,90.75)
            (G4,45.39)
            (G5,33.36)
        };   
        \addplot
        coordinates {
            (G1,83.51)
            (G2,79.98)
            (G3,83.84)
            (G4,52.44)
            (G5,43.24)
        };
        \legend{HGNN, UniGCN, MLP}
        \end{axis}
        \end{tikzpicture}
        
    \end{tabular}
    \vspace{-4mm}
    \caption{Observations. (a) Heterophilic neighbors follows a long-tail distribution and (b) the accuracy of HNN models drops significantly on highly heterophilic nodes in G4/G5.}\label{fig:observations}
    \vspace{-4mm}
\end{figure}

\vspace{1mm}
\noindent
\textbf{Observations.}
Figure~\ref{fig:observations} presents two key observations\footnote{Similar observations hold across the other datasets: Cora, Pubmed, and DBLP.} in Citeseer dataset~\cite{chien2021you}. 
First, \textbf{(a)} \textit{the number of heterophilic neighbors follows a \textit{long-tail} distribution}, 
indicating that a majority of nodes have only a few heterophilic neighbors, while a small number of nodes have many heterophilic neighbors. 
Second, \textbf{(b)} \textit{the accuracy of HNN teachers significantly decreases on highly heterophilic nodes}. 
Specifically, from the most homophilic group (G1) to the most heterophilic group (G5), both HGNN and UniGCN exhibit a substantial accuracy drop of approximately 40--50\%, 
eventually performing worse than a lightweight MLP that relies solely on node features without exploiting hypergraph structure. 
These observations suggest that teacher predictions are not equally reliable across all nodes and should be transferred discriminately during distillation.

\vspace{1mm}
\noindent
\textbf{Our work.}
Motivated by these observations, we propose a novel hypergraph distillation method, 
named \textbf{\underline{H}}eterophily-aware \textbf{\underline{AD}}aptiv\textbf{\underline{E}} di\textbf{\underline{S}}tillation (\textbf{{\method}}).
The key idea of {\method} is to consider node heterophily as a proxy for the reliability of teacher knowledge. 
To this end, {\method} first quantifies the heterophily of individual nodes and then adaptively regulates the amount of teacher knowledge transferred to a student model accordingly. 
Since {\method} focuses on the reliability of teacher knowledge rather than a specific distillation mechanism, 
it is \textit{agnostic} to both distillation methods and distilled knowledge (e.g., logits or embeddings), making it seamlessly applicable to existing hypergraph distillation methods.

\vspace{1mm}
\noindent
\textbf{Contributions.} The main contributions of this work are as follows.
\begin{itemize}[leftmargin=10pt]
    \item \textbf{Observations}: We observe that (i) node heterophily exhibits a long-tail distribution and (ii) HNN teachers suffer substantial performance degradation on highly heterophilic nodes, indicating that teacher knowledge is not equally reliable across all nodes.
    \item \textbf{Method}: We propose a novel hypergraph distillation method, \textbf{{\method}}, that quantifies node heterophily and adaptively transfers teacher knowledge accordingly.
    \item \textbf{Evaluation}: Extensive experiments using two HNN teacher models on four real-world hypergraphs demonstrate that {\method} consistently improves the accuracy of student models across diverse types of distilled knowledge, often outperforming even their teachers while achieving over 12.3$\times$ faster inference. 
\end{itemize}

\section{Related Work}\label{sec:related}
\noindent
\textbf{Hypergraph knowledge distillation.} 
Hypergraph knowledge distillation remains in its infancy, with only a handful of works having been studied.
LightHGNN~\cite{feng2024lighthgnn} distills knowledge from an HNN teacher to a lightweight MLP student through logits. 
LightHGNN+, an improved version of LightHGNN, injects noise into hyperedge features to estimate hyperedge reliability and selectively distills knowledge from hyperedges robust to perturbations.
DistillHGNN~\cite{forouzandeh2025distillhgnn} employs a light graph neural network model (i.e., TinyGCN~\cite{yan2020tinygnn}) as a student, which takes a clique-expanded hypergraph as input. 
DistillHGNN transfers knowledge from an HNN teacher through both logit distillation and contrastive learning.
ARCHER~\cite{xu2026adaptive} estimates node-level and hyperedge-level confidence of the teacher model and adaptively controls the strength of teacher guidance during knowledge distillation process.

\vspace{1mm}
\noindent
\textbf{Relation to our work.} 
Existing hypergraph distillation methods regulate teacher guidance based on prediction confidence or robustness to perturbation.
However, they overlook that HNN teachers are inherently less reliable on heterophilic nodes, 
leaving the relationship between node heterophily and the reliability of teacher knowledge unexplored.
In contrast, {\method} quantifies node heterophily as a proxy for teacher reliability and adaptively transfers teacher knowledge accordingly. 
Moreover, {\method} is agnostic to any distillation methods and can be readily integrated into them.

\section{Proposed Method: {\method}}\label{sec:proposed}

\subsection{Preliminaries}\label{sec:proposed-preliminary}
\noindent
\textbf{Notations}.
A hypergraph is defined as \( H = (V, E) \), where \( V = \{v_1, v_2, \dots, v_{|V|}\} \) and \( E = \{e_1, e_2, \dots, e_{|E|}\} \).
A hypergraph can generally be represented by an \textit{incidence} matrix $\mathbf{H}\in \{0,1\}^{|V|\times |E|}$,
where each element $h_{ij}=1$ if $v_i \in e_j$, and $h_{ij}=0$ otherwise.
The node and hyperedge feature matrices are denoted by \( \mathbf{P} \in \mathbb{R}^{|V| \times d} \), \( \mathbf{Q} \in \mathbb{R}^{|E| \times d} \), where each row \( p_i \) and \( q_i \) represents the \( d\)-dimensional feature vector of a node and a hyperedge, respectively.

\vspace{1mm}
\noindent
\textbf{Hypergraph neural networks.}  
Given a hypergraph $H = (V, E)$, 
a hypergraph neural network (HNN) model produces $\mathbf{P} \in \mathbb{R}^{|V| \times d}$ and $\mathbf{Q} \in \mathbb{R}^{|E| \times d}$.
This process repeats two steps:
(1) (\textit{node-to-hyperedge}) producing a hyperedge embedding by aggregating the node embeddings and (2) (\textit{hyperedge-to-node}) producing a node embedding by aggregating the hyperedge embeddings.
Formally, the node and hyperedge embeddings at the \textit{l}-th layer are defined as:
\begin{align}
    \footnotesize \mathbf{Q}^{(l)} = \sigma \left(\mathbf{H}^T \mathbf{P}^{(l-1)} \mathbf{W}^{(l)}_E + b^{(l)}_E \right), 
    \footnotesize \mathbf{P}^{(l)} = \sigma \left(\mathbf{H} \mathbf{Q}^{(l)} \mathbf{W}^{(l)}_V+ b^{(l)}_V 
    \right),
    \label{eq:hypergnn}
\end{align}
where $\mathbf{P}^{(0)}=\mathbf{X}$, $\mathbf{W}^{(l)}_*$ and $b^{(l)}_*$ are trainable weight and bias matrices,
and $\sigma$ is an activation function.

\subsection{Heterophily Quantification}\label{sec:proposed-hetero-quantify}
As observed in Section~\ref{sec:intro}, 
HNN teachers exhibit substantially lower accuracy on highly heterophilic nodes. 
Motivated by this observation, {\method} quantifies node heterophily and uses it as a proxy for the reliability of teacher knowledge.
A straightforward way to quantify node heterophily is to count the number of \textit{heterophilic neighbors}, i.e., nodes connected through hyperedges that have different class labels. 
Intuitively, a node connected to many different-label neighbors is more likely to receive conflicting relational signals and thus may be more difficult for an HNN teacher to model.

However, such a formulation does not fully capture the structural characteristics of hypergraphs. 
For example, consider two nodes that have 10 different-label neighbors. 
The first node participates in 10 hyperedges, each containing only one different-label neighbor, 
whereas the second node participates in a single hyperedge containing all 10 different-label neighbors. 
Although both nodes have the same number of heterophilic neighbors, the second node is intuitively more heterophilic because it is involved in a hyperedge with substantially higher label diversity.

To capture such characteristics of hypergraphs, 
we quantify node heterophily through the heterophily of its incident hyperedges. 
Specifically, we first \textbf{(a)} estimate the heterophily of each hyperedge based on the entropy of its member-node labels and then \textbf{(b)} aggregate the estimated heterophily scores of incident hyperedges to obtain the heterophily score of a node.

\vspace{1mm}
\noindent \textbf{(a) Hyperedge heterophily.}
Formally, the heterophily of a hyperedge is defined as the entropy of its member-node labels:
\begin{equation}
    h(e)= -\sum_{c \in \mathcal{C}} p_c \log p_c,\label{eq:hyperedge-hetero}
\end{equation}
where $\mathcal{C}$ denotes the set of class labels and $p_c$ is the proportion of nodes in hyperedge $e$ belonging to class $c$. 
Thus, hyperedges containing diverse classes yield high heterophily scores.

\noindent \textbf{(b) Node heterophily.}
The heterophily of a node is estimated by aggregating the heterophily scores of its incident hyperedges:
\begin{equation}
    h(v)=\sum_{e \in \mathcal{E}(v)} w_e\cdot h(e),\label{eq:node-hetero}
\end{equation}
where $\mathcal{E}(v)$ denotes the set of hyperedges incident to node $v$, and $w_e$ determines the contribution of hyperedge $e$.

We consider four weighting strategies for $w_e$, where all weights are normalized such that $\sum_{e\in\mathcal{E}(v)} w_e = 1$.
\begin{itemize}[leftmargin=10pt]
    \item (1) \textit{Uniform}: $w_e=\frac{1}{|\mathcal{E}(v)|}$, which assigns equal importance to all incident hyperedges.
    \item (2) \textit{Weighted}: $w_e= \frac{|e|}{\sum_{e' \in \mathcal{E}(v)} |e'|}$, which assigns larger weights to larger hyperedges under the assumption that they contribute more strongly to node heterophily.
    \item (3) \textit{Bound-Weighted}: $w_e=\frac{\min(|e|,Q_3)}{\sum_{e' \in \mathcal{E}(v)}\min(|e'|,Q_3)}$, where $Q_3$ denotes the third quartile of the hyperedge-size distribution. This strategy limits the influence of extremely large hyperedges while preserving the effect of hyperedge size.
    \item (4) \textit{Log-Weighted}: $w_e= \frac{\log(|e|+1)} {\sum_{e' \in \mathcal{E}(v)}\log(|e'|+1)}$, which reduces the dominance of large hyperedges by applying logarithmic scaling.
\end{itemize}

We use the uniform strategy as the default setting of {\method} and compare all weighting strategies in EQ2 of Section~\ref{sec:exp-results}.

\subsection{Heterophily-Aware Distillation}\label{sec:proposed-distillation}
While {\method} can be applied to any hypergraph tasks, we focus on the node classification task.
Given a pretrained HNN teacher and a lightweight MLP student, both models take node features as input and produce predictions for target nodes.

Formally, given the input node feature matrix $\mathbf{P}^{(0)}$, 
the teacher model $\theta^T$ and the student model $\theta^S$ generate node representations, $\mathbf{P}^{T} \leftarrow \theta^{T}(\mathbf{P}^{(0)})$ and $\mathbf{P}^{S} \leftarrow \theta^{S}(\mathbf{P}^{(0)})$, respectively. 
The resulting representations are then fed into prediction layers to produce the teacher and student predictions,
$\mathbf{Z}^{T} \leftarrow f^{T}(\mathbf{P}^{T})$ and $\mathbf{Z}^{S} \leftarrow f^{S}(\mathbf{P}^{S})$,
where $\mathbf{Z}^{T}, \mathbf{Z}^{S}\in\mathbb{R}^{|V|\times C}$ denote the teacher and student logits for $|V|$ target nodes and $C$ classes, respectively.
Then, the student model $\theta^S$ is trained based on the following loss function:
\begin{equation}
\mathcal{L} = \mathcal{L}_{CE} + \alpha\cdot \mathcal{L}_{Distill}, \label{eq:student-loss}
\end{equation}
where $\mathcal{L}_{CE}$ is the cross-entropy loss between student predictions and ground-truth labels, $\mathcal{L}_{Distill}$ is the distillation loss between student and teacher predictions/embeddings,
and $\alpha$ is the hyperparameter to controls the strength of distillation loss.

We consider three types of knowledge distillation: 
logit distillation (L), embedding distillation (E), and their joint distillation (L+E). 
Thus, $\mathcal{L}_{Distill}$ is generally defined as follows:
\begin{equation}
\mathcal{L}_{Distill} = \lambda_{L}\cdot\mathcal{L}_{L} + \lambda_{E}\cdot\mathcal{L}_{E},
\end{equation}
where $\mathcal{L}_{L}$ denotes the logit distillation loss and $\mathcal{L}_{E}$ denotes the embedding distillation loss. 
Specifically, $\mathcal{L}_{L}$ is defined as the KL divergence between teacher and student predictions~\cite{hinton2015distilling}, 
while $\mathcal{L}_{E}$ is defined as the mean squared error (MSE) between teacher and student node representations~\cite{zheng2022cold}. 
The coefficients $\lambda_L,\lambda_E\in{[0,1]}$ determine the distillation objective. 
Specifically, $(\lambda_L,\lambda_E)=(\lambda_L,0)$, $(0,\lambda_E)$, and $(\lambda_L,\lambda_E)$ correspond to L, E, and L+E, respectively.

To adaptively transfer teacher knowledge, {\method} assigns a reliability weight to each node based on its heterophily score. 
Specifically, for each node $v$, the reliability weight is defined as
\begin{equation}
r(v) = \exp(-\beta h(v)),\label{eq:reliability}
\end{equation}
where $h(v)$ is the node heterophily score and $\beta$ is the hyperparameter to control how strongly node heterophily affects the reliability weight. 
Thus, homophilic nodes with low heterophily receive large reliability weights, while heterophilic nodes with high heterophily receive small weights.

Finally, {\method} adaptively weights the distillation loss of each node as follows:
\begin{equation}
\mathcal{L}^{*}_{Distill} =
\frac{\sum_{v\in V} r(v) \cdot \left( \lambda_{L}\mathcal{L}_{L}(v) + \lambda_{E}\mathcal{L}_{E}(v) \right)}{\sum_{v'\in V} r(v')},
\end{equation}
where $\mathcal{L}_{L}(v)$ and $\mathcal{L}_{E}(v)$ denote the logit and embedding distillation losses of node $v$, respectively.
As a result, homophilic nodes with reliable teacher knowledge contribute more to the distillation process, whereas heterophilic nodes with potentially unreliable teacher knowledge contribute less.

\section{Experimental Validation}\label{sec:experiment}
In this section, we evaluate comprehensively {\method} by answering the following evaluation questions (EQs): 
\begin{itemize}[leftmargin=8pt]
    \item \textbf{EQ1} (\textit{Distillation performance}). To what extent does {\method} improve the performance of distilled student models?
    \item \textbf{EQ2} (\textit{Heterophily quantification}). How effective is our heterophily quantification strategy in enhancing distillation performance?
    \item \textbf{EQ3} (\textit{Sensitivity analysis}). How sensitive is the performance of {\method} to the hyperparameter $\beta$? 
\end{itemize}

\subsection{Experimental Setup}\label{sec:exp-setup}

\noindent
\textbf{Datasets and HNN models.}
We use four real-world hypergraphs (see Table~\ref{table:datasets} for details): 
(1) three co-citation datasets (Citeseer, Cora, and Pubmed) and (2) one authorship dataset (DBLP-A).
In the co-citation datasets, each node represents a paper and each hyperedge represents a group of papers co-cited by a paper;
in the authorship dataset, each node represents a paper and each hyperedge represents a group of papers written by an author.
For all the datasets, we use the bag-of-word features from the abstract of each paper as in~\cite{ko2023cash,yu2025hygen}.
We select two hypergraph neural network models as teacher models in our experiments (HGNN~\cite{feng2019hypergraph} and UniGCN~\cite{huang2021unignn}).

\vspace{1mm}
\noindent
\textbf{Task and evaluation.}
We consider the node classification task that aims to predict the class labels of unseen nodes. 
Following~\cite{chien2021you}, 
nodes are split into training (50\%), validation (25\%), and test (25\%) sets.
We (1) measure the accuracy on the test set when the accuracy on the validation set is maximized, 
and (2) report the averaged test accuracy over five runs with different random seeds.

\begin{table}[h]
\centering
\small
\caption{Statistics of real-world hypergraphs.}
\vspace{-4mm}
\label{table:datasets}
\setlength\tabcolsep{4pt}
\def\arraystretch{0.8} 
\begin{tabular}{l|rrrcc}
\toprule

\multirow{2}{*}{Datasets} & \multirow{2}{*}{$|\mathcal{V}|$} & \multirow{2}{*}{$|\mathcal{E}|$} & \multirow{2}{*}{$|\mathcal{C}|$} & \multirow{2}{*}{$|\mathcal{F}|$} & Hyperedge size dist. \\
&  &  &  &  & (min/Q1/Q2/Q3/max) \\
\midrule

 \textbf{Citeseer} & 1,458 & 1,079 & 6 & 3,703 & 2/2/2/4/26  \\ 
 \textbf{Cora} & 1,434 & 1,579 & 7 & 1,433 & 2/2/3/4/5  \\ 
 \textbf{Pubmed} & 3,840 & 7,963 & 3 & 500 & 2/2/3/4/171  \\ 
 \textbf{DBLP} & 41,302 & 22,363 & 6 & 4,543 & 2/2/3/5/202  \\ 

\bottomrule
\end{tabular}
\end{table}

\begin{table*}[t]
\centering
\small
\caption{Hypergraph Distillation Performance: {\method} consistently improves the accuracy of distilled student models across two teacher models, four datasets, and three types of distilled knowledge, compared to the baseline.}
\label{tab:eq1}
\vspace{-3mm}
\setlength\tabcolsep{3.5pt} 
\def\arraystretch{0.75} 

\begin{tabular}{cc|cc cc cc cc}
\toprule

 \multirow{3}{*}{Distilled} & Datasets & \multicolumn{2}{c}{\textbf{Citeseer}} & \multicolumn{2}{c}{\textbf{Cora}} & \multicolumn{2}{c}{\textbf{Pubmed}} & \multicolumn{2}{c}{\textbf{DBLP}}\\

 \cmidrule(lr){3-4} \cmidrule(lr){5-6} \cmidrule(lr){7-8} \cmidrule(lr){9-10}

 & \multirow{2}{*}{Teacher} & T1: HGNN & T2: UniGCN & T1: HGNN & T2: UniGCN & T1: HGNN & T2: UniGCN & T1: HGNN & T2: UniGCN \\
 &  & (0.733) & (0.721) & (0.832) & (0.825) & (0.849) & (0.852) & (0.906) & (0.905)  \\

\midrule
\multirow{3}{*}{\textbf{L}}
 & Baseline & 0.744 $\pm$ 0.014 & 0.730 $\pm$ 0.010 & 0.839 $\pm$ 0.012 & 0.826 $\pm$ 0.018 & 0.866 $\pm$ 0.008 & 0.859 $\pm$ 0.010 & 0.914 $\pm$ 0.003 & 0.915 $\pm$ 0.002 \\
 & \textbf{{\method}} & \textbf{0.753} $\pm$ 0.019 & \textbf{0.747} $\pm$ 0.014 & \textbf{0.858} $\pm$ 0.005 & \textbf{0.836} $\pm$ 0.014 & \textbf{0.886} $\pm$ 0.005 & \textbf{0.878} $\pm$ 0.007 & \textbf{0.923} $\pm$ 0.003 & \textbf{0.925} $\pm$ 0.003 \\
 \cmidrule(lr){2-10}
 & Gain & \textcolor{blue}{+1.21\%} & \textcolor{blue}{+2.33\%} & \textcolor{blue}{+2.26\%} & \textcolor{blue}{+1.21\%} & \textcolor{blue}{+2.31\%} & \textcolor{blue}{+2.21\%} & \textcolor{blue}{+2.11\%} & \textcolor{blue}{+1.10\%}  \\
 
\midrule

\multirow{3}{*}{\textbf{E}}
 & Baseline & 0.736 $\pm$ 0.020 & 0.729 $\pm$ 0.019 & 0.831 $\pm$ 0.005 & 0.767 $\pm$ 0.085 & 0.860 $\pm$ 0.007 & 0.803 $\pm$ 0.019 & 0.873 $\pm$ 0.005 & 0.855 $\pm$ 0.005 \\
 & \textbf{{\method}} & \textbf{0.747} $\pm$ 0.015 & \textbf{0.750} $\pm$ 0.018 & \textbf{0.836} $\pm$ 0.012 & \textbf{0.826} $\pm$ 0.020 & \textbf{0.871} $\pm$ 0.007 & \textbf{0.815} $\pm$ 0.013 & \textbf{0.873} $\pm$ 0.005 & \textbf{0.855} $\pm$ 0.003 \\
 \cmidrule(lr){2-10}
 & Gain & \textcolor{blue}{+1.49\%} & \textcolor{blue}{+2.88\%} & \textcolor{blue}{+0.60\%} & \textcolor{blue}{+7.69\%} & \textcolor{blue}{+1.28\%} & \textcolor{blue}{+1.49\%} & +0.00\% & +0.00\%  \\

 \midrule

 \multirow{3}{*}{\textbf{L+E}}
 & Baseline & 0.745 $\pm$ 0.019 & 0.733 $\pm$ 0.010 & 0.827 $\pm$ 0.016 & 0.815 $\pm$ 0.018 & 0.861 $\pm$ 0.006 & 0.868 $\pm$ 0.007 & 0.913 $\pm$ 0.003 & 0.914 $\pm$ 0.003 \\
 & \textbf{{\method}} & \textbf{0.752} $\pm$ 0.017 & \textbf{0.740} $\pm$ 0.012 & \textbf{0.863} $\pm$ 0.005 & \textbf{0.836} $\pm$ 0.011 & \textbf{0.882} $\pm$ 0.003 & \textbf{0.875} $\pm$ 0.005 & \textbf{0.923} $\pm$ 0.003 & \textbf{0.923} $\pm$ 0.002 \\
 \cmidrule(lr){2-10}
 & Gain & \textcolor{blue}{+0.94\%} & \textcolor{blue}{+0.95\%} & \textcolor{blue}{+4.35\%} & \textcolor{blue}{+2.58\%} & \textcolor{blue}{+2.44\%} & \textcolor{blue}{+0.81\%} & \textcolor{blue}{+1.10\%} & \textcolor{blue}{+0.98\%}  \\

\bottomrule
\end{tabular}
\end{table*}

\subsection{Experimental Results}\label{sec:exp-results}
\noindent
\textbf{EQ1: Distillation performance}.
In this experiment, we evaluate {\method} with three types of distilled knowledge: 
logit distillation (L), embedding distillation (E), and their joint distillation (L+E).
For each knowledge type, we compare {\method} with a baseline that directly distills teacher knowledge without considering its reliability.
Table~\ref{tab:eq1} shows the node classification accuracy of student models distilled from two HNN teachers. 
{\method} consistently improves the student accuracy across all datasets, teacher models, and types of distilled knowledge, 
achieving gains of up to 7.69\% over the baseline. 
Notably, under L and L+E distillation, the distilled students even outperform their teachers across all datasets and teachers. 
These results suggest that teacher knowledge is not always reliable across all nodes and should be transferred discriminately. 
Therefore, by adaptively transferring teacher knowledge based on node heterophily, 
{\method} achieves more effective knowledge distillation.

Table~\ref{tab:inference} shows the inference time of two teachers (3-layer HGNN and UniGCN) and a student (3-layer MLP). 
Clearly, the student achieves significantly faster inference than their teachers, 
up to $12.3\times$ speedup on Pubmed.
The speedup tends to become more pronounced on larger hypergraphs because the computational cost of HNNs grow exponentially with the increasing size of hypergraphs.

\begin{table}[t]
\centering
\small
\caption{Inference time of two teacher models and {\method}.}\label{tab:inference}
\vspace{-4mm}
\setlength\tabcolsep{3pt} 
\def\arraystretch{0.8} 
\begin{tabular}{c|cccc}
\toprule
 & \textbf{Citeseer} & \textbf{Cora} & \textbf{Pubmed} & \textbf{DBLP} \\
\midrule
T1: HGNN & 0.56 (ms) & 0.54 (ms)& 0.94 (ms)& 5.27 (ms) \\
T2: UniGCN & 1.16 (ms) & 0.58 (ms)& 1.84 (ms)& 14.70 (ms) \\
\midrule

\multirow{2}{*}{\textbf{{\method}}} & \textbf{0.18} (ms) & \textbf{0.12} (ms) & \textbf{0.15} (ms) & \textbf{1.50} (ms) \\
 & (\textcolor{blue}{$3.1\times$ / $6.4\times$}) & (\textcolor{blue}{$4.5\times$ / $4.8\times$})  & (\textcolor{blue}{$6.3\times$ / $12.3\times$})  & (\textcolor{blue}{$3.5\times$ / $9.8\times$}) \\

\bottomrule
\end{tabular}
\end{table}

\begin{table}[t]
\centering
\small
\caption{Effect of heterophily quantification strategies.}\label{tab:eq2}
\vspace{-4mm}
\setlength\tabcolsep{2.9pt} 
\def\arraystretch{0.8} 
\begin{tabular}{cl|cccc|c}
\toprule
Distilled & Method & \textbf{Citeseer} & \textbf{Cora} & \textbf{Pubmed} & \textbf{DBLP} & Rank $\downarrow$ \\
\midrule

\multirow{5}{*}{\textbf{L}}
& Baseline             & 0.744 & 0.839 & 0.866 & 0.914 & 5.00 \\
 \cmidrule(lr){2-7}
& Uniform                 & \textbf{0.753} & 0.858 & \textbf{0.886} & \textbf{0.923} & \textbf{1.75} \\
& Weighted         & 0.748 & 0.861 & 0.880 & 0.922 & 3.00 \\
& Bound-Weighted & 0.751 & 0.859 & 0.882 & 0.922 & 2.25 \\
& Log-Weighted     & 0.750 & \textbf{0.864} & 0.882 & 0.919 & 2.50 \\
\midrule

\multirow{5}{*}{\textbf{E}}
& Baseline             & 0.736 & 0.831 & 0.860 & 0.873 & 3.25 \\
\cmidrule(lr){2-7}
& Uniform                  & 0.747 & 0.836 & \textbf{0.871} & \textbf{0.873} & \textbf{2.25} \\
& Weighted         & 0.746 & \textbf{0.849} & 0.817 & 0.859 & 3.75 \\
& Bound-Weighted  & \textbf{0.752} & 0.847 & 0.842 & 0.860 & 2.50 \\
& Log-Weighted     & 0.751 & 0.847 & 0.828 & 0.863 & 2.75 \\
\midrule

\multirow{5}{*}{\textbf{L+E}}
& Baseline             & 0.745 & 0.827 & 0.861 & 0.913 & 5.00 \\
\cmidrule(lr){2-7}
& Uniform                  & \textbf{0.752} & 0.863 & \textbf{0.882} & \textbf{0.923} & 1.75 \\
& Weighted         & 0.751 & 0.863 & 0.876 & 0.922 & 3.25 \\
& Bound-Weighted  & \textbf{0.752} & \textbf{0.864} & 0.879 & \textbf{0.923} & \textbf{1.50} \\
& Log-Weighted      & \textbf{0.752} & 0.861 & 0.880 & \textbf{0.923} & 2.00 \\

\bottomrule
\end{tabular}
\end{table}

\vspace{1mm}
\noindent
\textbf{EQ2: Heterophily quantification.}
In this experiment, we compare four different strategies for quantifying node heterophily using HGNN as the teacher model and logit distillation as the distillation objective.
As shown in Table~\ref{tab:eq2}, \textit{all} strategies outperform the baseline on average.
Notably, among the strategies, Bound-Weighted generally achieves the second-best performance (best performer on L+E). 
This result suggests that not all hyperedges contribute equally to node heterophily. 
In particular, larger hyperedges tend to have a greater influence on node heterophily, and incorporating hyperedge size enables more accurate heterophily estimation and, consequently, more effective knowledge transfer.

\begin{figure}
    \centering
    \begin{tabular}{cc}
    \hspace{-5mm}
        \begin{tikzpicture}
        \begin{axis}[
            width=0.53\linewidth,
        	height=3.2cm,
            grid=both,
            major grid style={line width=.2pt,draw=gray!50},
            ylabel={Accuracy},
            xlabel style={yshift=9pt,font=\scriptsize},
            ylabel style={yshift=-15pt,font=\scriptsize},
            xticklabel style={font=\scriptsize},
            yticklabel style={font=\scriptsize},
            xlabel={Heterophily control factor $\beta$},
            xmin=0.5, xmax=6.5,
            ymin=0.73, ymax=0.76,
            xtick={1, 2, 3, 4, 5, 6},
            ytick={0.725, 0.73, 0.735, 0.74, 0.745, 0.75, 0.755, 0.76},
            yticklabels={0.725, 0.73, , 0.74, , 0.75, , 0.76},
            every axis plot/.append style={thick}
            ]
            \addplot[
                mark=triangle,
                color=blue,
                line width=0.7pt,
                solid
                ]table[x=beta, y=accuracy]{sensitivity-citeseer.txt};
            
            \addplot [
                red!80, no markers, dashed,
            ] coordinates {(0,0.744) (7,0.744)}; 
        \end{axis}
        \end{tikzpicture}
        &  
        \hspace{-3mm}
        \begin{tikzpicture}
        \begin{axis}[
            width=0.53\linewidth,
        	height=3.2cm,
            grid=both,
            major grid style={line width=.2pt,draw=gray!50},
            ylabel={Accuracy},
            xlabel style={yshift=9pt,font=\scriptsize},
            ylabel style={yshift=-15pt, font=\scriptsize},
            xticklabel style={font=\scriptsize},
            yticklabel style={font=\scriptsize},
            xlabel={Heterophily control factor $\beta$},
            xmin=0.5, xmax=6.5,
            ymin=0.825, ymax=0.875,
            xtick={1, 2, 3, 4, 5, 6},
            ytick={0.825, 0.8375, 0.85, 0.8625, 0.875, 0.8875, 0.90},
            every axis plot/.append style={thick}
            ]
            \addplot[
                mark=triangle,
                color=blue,
                line width=0.7pt,
                solid
                ]
            table[x=beta, y=accuracy]{sensitivity-cora.txt};
            
            \addplot [
                red!80, no markers, dashed,
            ] coordinates {(0,0.839) (7,0.839)}; 
        
        \end{axis}
        \end{tikzpicture}
        \vspace{-1mm}
         \\
         (a) Citeseer & (b) Cora \\
    \end{tabular}
    \vspace{-4mm}
    \caption{Sensitivity analysis. {\method} consistently outperforms the baseline across all values of the hyperparameter $\beta$.}
    \vspace{-4mm}
    \label{fig:eq3-sensitivity}
\end{figure}

\vspace{1mm}
\noindent
\textbf{EQ3: Sensitivity Analysis.}
Finally, we evaluate the sensitivity of the performance of {\method} to the hyperparameter $\beta$.
Figure~\ref{fig:eq3-sensitivity} shows the performance of {\method} using HGNN teacher with logit distillation. 
Across a wide range of $\beta$, {\method} outperforms the baseline (red dotted line) that does not consider node heterophily. 
In general, the performance tends to improve as $\beta$ increases, indicating that suppressing potentially unreliable teacher knowledge from highly heterophilic nodes is beneficial. 
The best performance is achieved around $\beta=4\sim5$ on both datasets. 
These results demonstrate that {\method} is insensitive to the choice of $\beta$ while consistently improving over the baseline.

\section{Conclusion}\label{sec:con}
In this paper, we observe that HNN teachers suffer from heterophilic nodes during distillation. 
Motivated by this observation, we quantify node heterophily as a proxy for teacher reliability and propose \textbf{{\method}}, a heterophily-aware adaptive hypergraph distillation method that selectively transfers teacher knowledge according to the estimated node heterophily.
Extensive experiments demonstrate that {\method} consistently improves the accuracy of student models across diverse hypergraphs, teacher models, and types of distilled knowledge, 
often outperforming even their HNN teachers while achieving up to $12.3\times$ faster inference.






\bibliographystyle{ACM-Reference-Format}
\bibliography{bibliography}


\end{document}